\documentclass[10pt,twocolumn,letterpaper]{article}
\usepackage[margin=0.75in,columnsep=0.3in]{geometry}
\usepackage[T1]{fontenc}
\usepackage{times}
\usepackage{amsmath,amssymb,bm,amsthm}
\usepackage{graphicx}
\usepackage{booktabs}
\usepackage{array}
\usepackage{subcaption}
\usepackage[table]{xcolor}  
\usepackage{tikz}
\usetikzlibrary{arrows.meta,positioning,calc}
\usepackage[numbers,sort&compress]{natbib}
\usepackage[colorlinks=true,linkcolor=blue!60!black,citecolor=blue!60!black,urlcolor=blue!60!black]{hyperref}
\usepackage[capitalize,noabbrev]{cleveref}

\newcommand{\G}{\mathcal{G}}                  
\newcommand{\Hm}{\mathcal{H}}                 
\newcommand{\src}{\mathbf{x}^{\mathrm{src}}}  
\newcommand{\tgt}{\mathbf{x}^{\mathrm{gt}}}   
\newcommand{\ours}{\textsc{HintErase}}

\newcommand{\ie}{i.e.,\ }
\newtheorem{proposition}{Proposition}
\def\reportauthors{Dongyang Zhang, Wei Zhang, Renjie Zou, and Zhiwei Huang}
\newcommand{\projecturl}{https://zdyshine.github.io/HintErase}
\newcommand{\pagecase}[1]{\href{\projecturl/\##1}{\texttt{\##1}}}  
\newcommand{\cell}[2][0.5625]{\IfFileExists{figures/img/#2.jpg}%
  {\includegraphics[width=\linewidth]{figures/img/#2.jpg}}%
  {\fcolorbox{black!25}{black!4}{\parbox[b][#1\linewidth][c]{\dimexpr\linewidth-2\fboxsep-2\fboxrule}%
    {\centering\tiny\color{black!45}\detokenize{#2}}}}}
\newcommand{\collab}[1]{{\scriptsize\sffamily #1}}
\newcommand{\rowlab}[1]{\rotatebox{90}{\scriptsize\sffamily #1}}

\hypersetup{
  pdftitle={Hint, Not Hole: Localization-Robust Video Overlay Removal},
  pdfauthor={\reportauthors},
  pdfsubject={Video overlay removal},
  pdfkeywords={video overlay removal; video inpainting; subtitle removal; diffusion transformer}
}
\title{Hint, Not Hole: Localization-Robust Video Overlay Removal\\[4pt]\large Technical Report}
\author{%
  \ifx\reportauthors\empty\else\reportauthors\\\fi
  Malanshan Audio \& Video Laboratory\\
  {\small \texttt{zhangdongyang@mlslabs.com.cn}}\\
  {\small Project page: \url{\projecturl}}
}
\date{}

\begin{document}
\twocolumn[{%
  \renewcommand\twocolumn[1][]{#1}%
  \maketitle
\begin{center}
\vspace{-6pt}
\setlength{\tabcolsep}{1pt}\renewcommand{\arraystretch}{0.6}
\begin{tabular}{@{}c p{0.268\textwidth} p{0.201\textwidth} p{0.206\textwidth} p{0.268\textwidth}@{}}
 & \centering\collab{Karaoke lyrics} & \centering\collab{Title card and label logo} & \centering\collab{Song lyrics in animation} & \centering\arraybackslash\collab{End credits in animation} \\
\rowlab{Input} &
 \cell{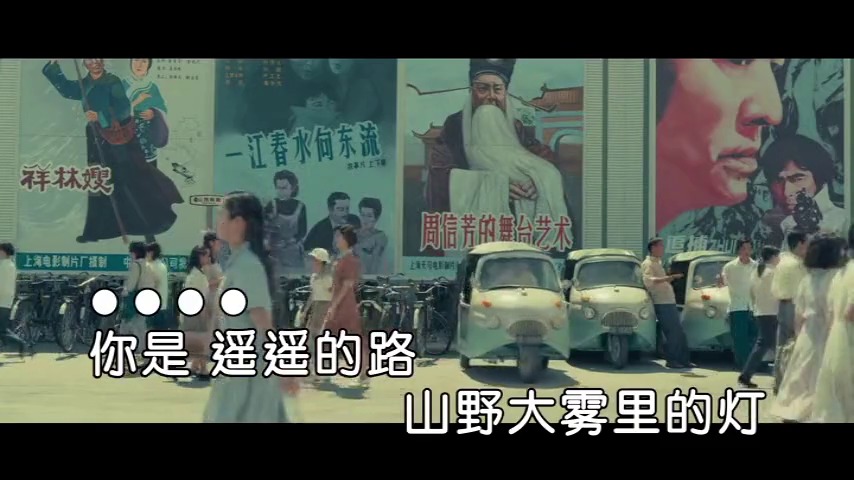} & \cell[0.75]{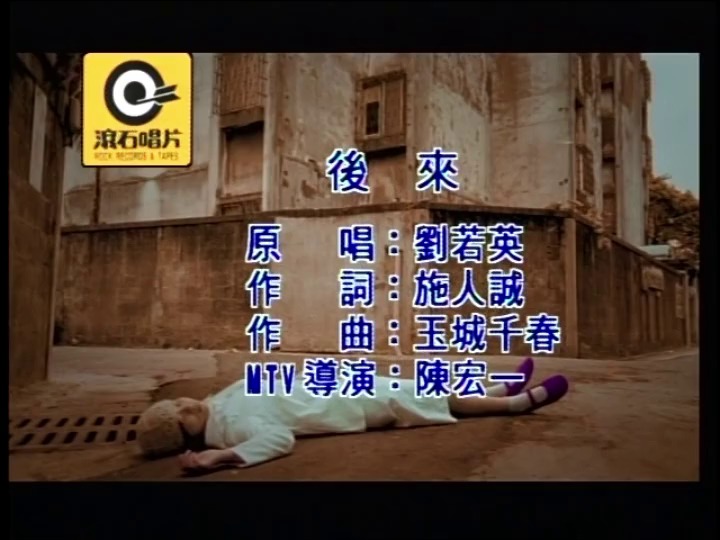} & \cell[0.731]{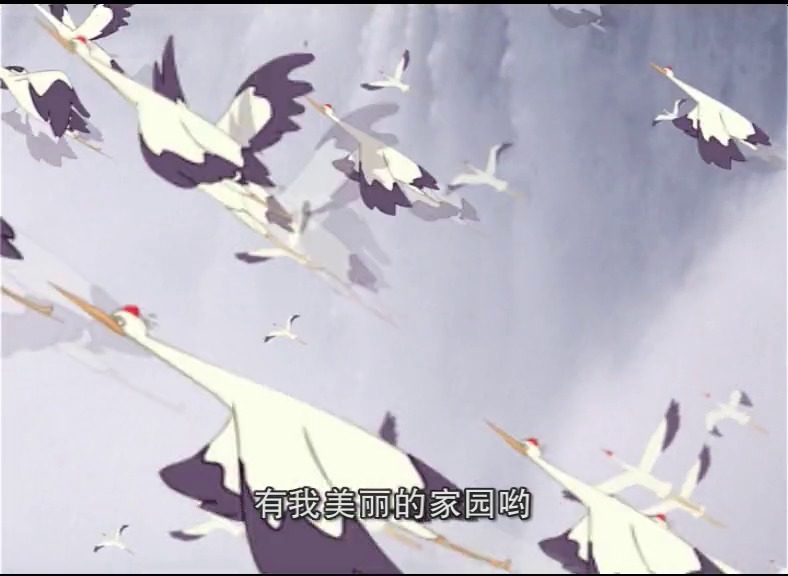} & \cell{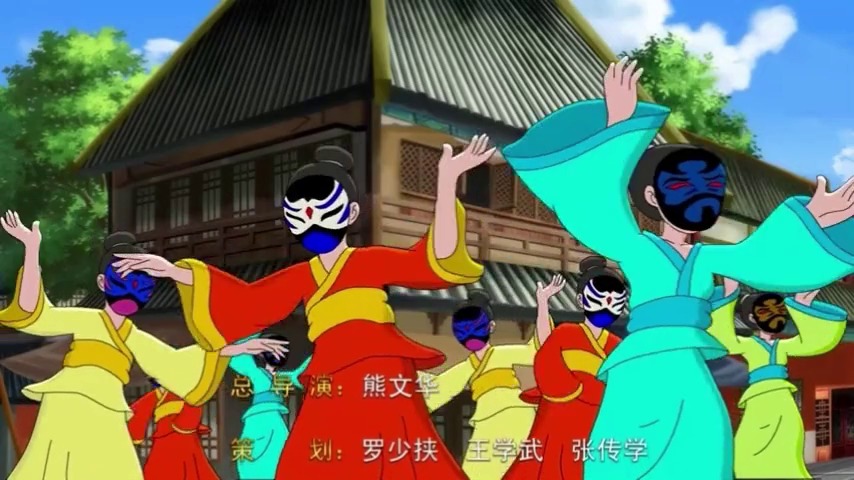} \\
\rowlab{Hint} &
 \cell{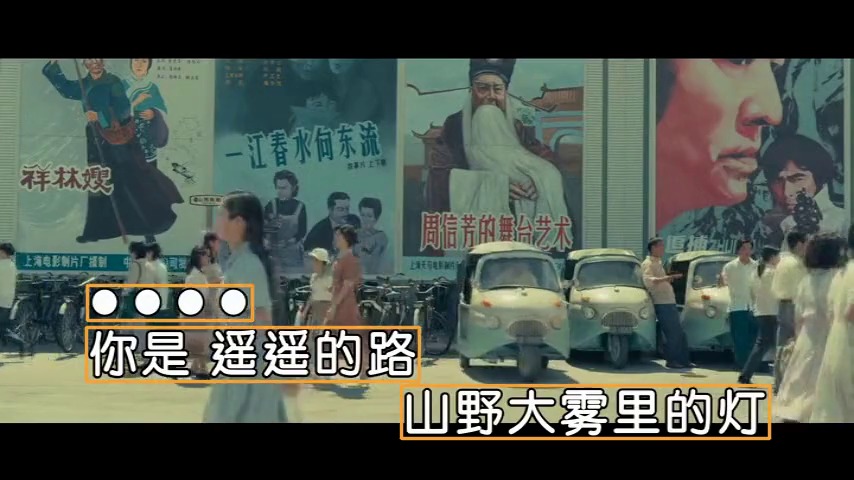} & \cell[0.75]{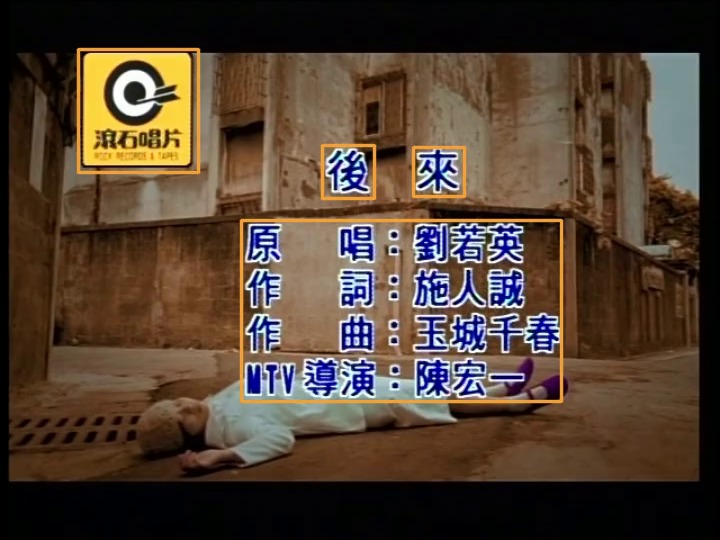} & \cell[0.731]{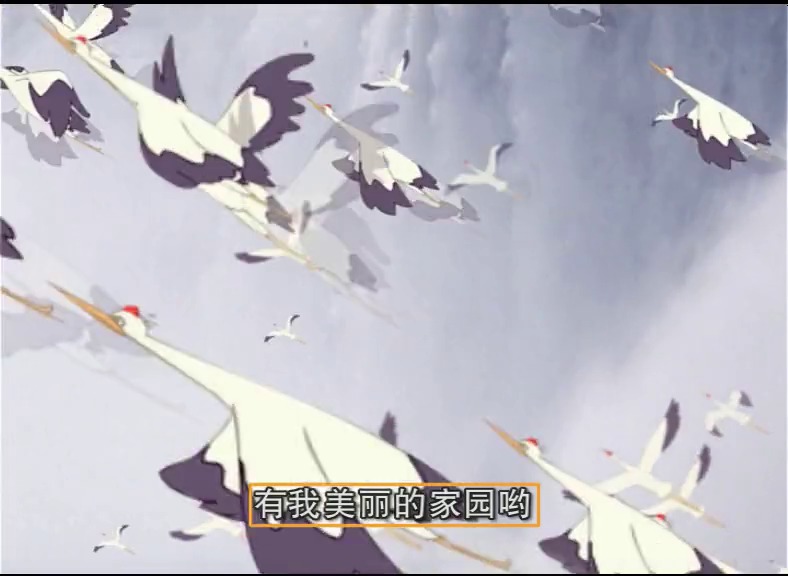} & \cell{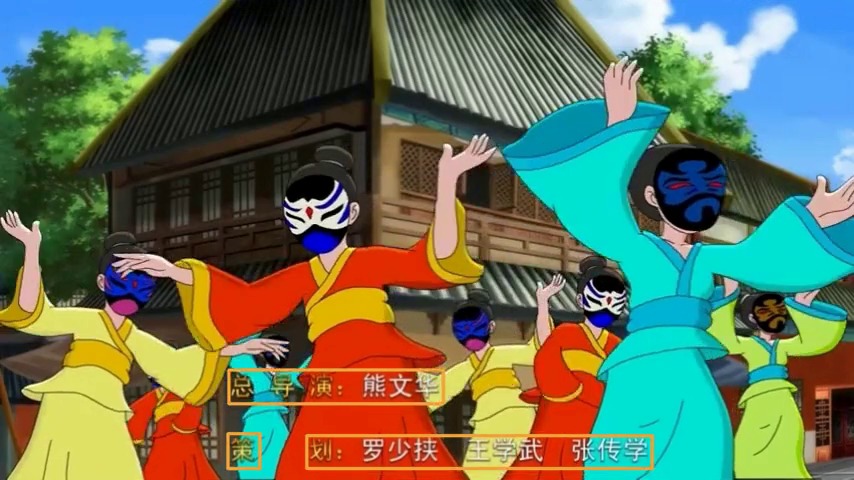} \\
\rowlab{\ours} &
 \cell{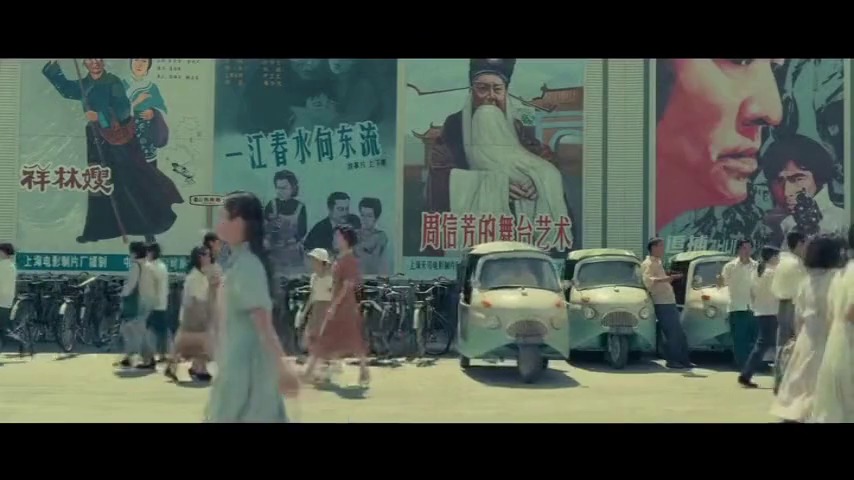} & \cell[0.75]{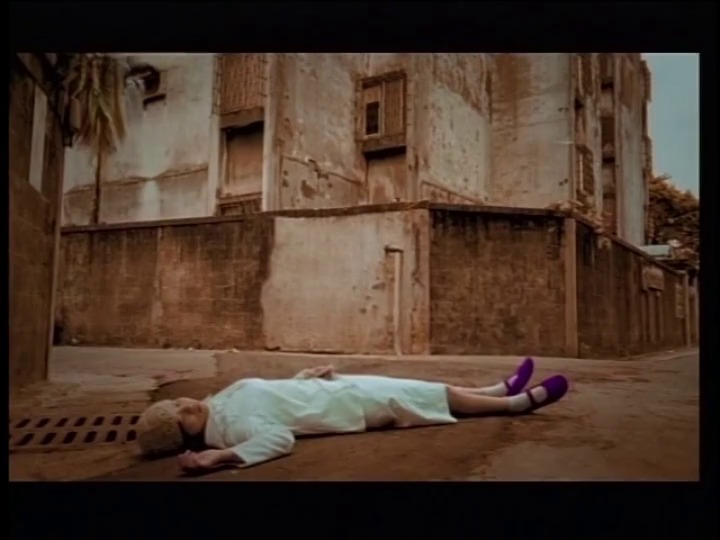} & \cell[0.731]{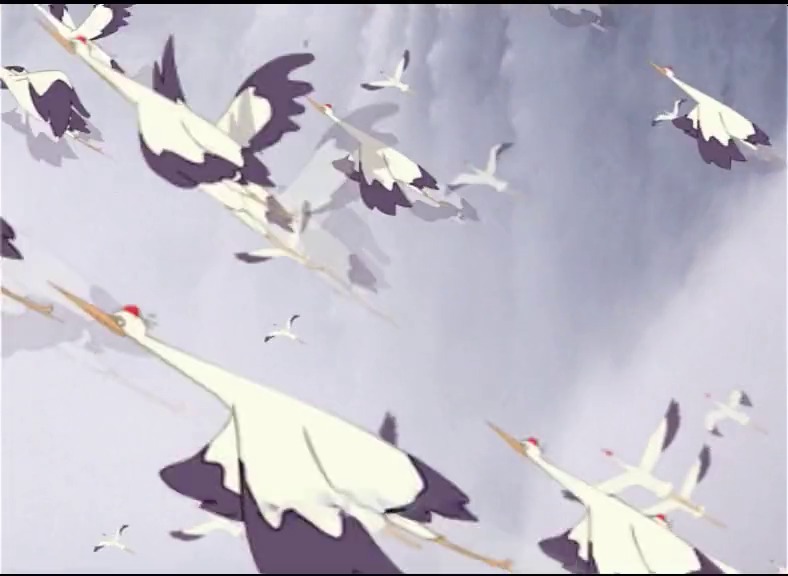} & \cell{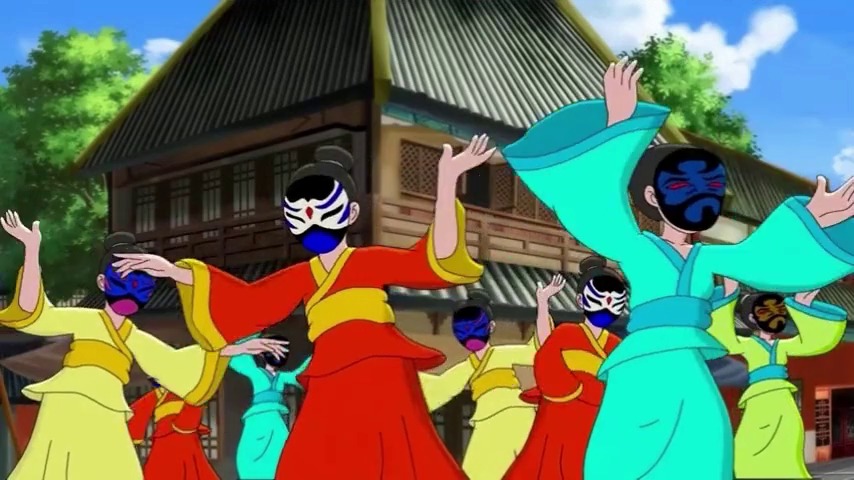} \\
\end{tabular}
\captionof{figure}{\textbf{One model for the overlays in a release.} \ours{} removes karaoke lyrics with countdown dots over a crowded street, a multi-line title card together with the label logo, theme-song lyrics over fast animation, and credit lines in mixed fonts, from standard-definition sources upward. The hint (orange) is one box per text block or logo; the model sees the whole frame with the overlay still in place. All four clips play on the project page (\pagecase{ktv-lyrics}, \pagecase{ktv-title}, \pagecase{anime-lyrics}, \pagecase{anime-credits}).}
\label{fig:teaser}
\end{center}
\vspace{4pt}

}]

\begin{abstract}
Video overlay removal recovers the scene beneath burned-in subtitles, karaoke lyrics, titles, credits, logos and watermarks.
Generative video models now fill a known region convincingly, but in practice the region comes from a detector or a segmenter, and it is wrong in both directions: it misses strokes, halos, whole frames and whole objects, and it covers background between characters, around boxes and on genuine scene text.
Erasers that treat the mask as a hole, regenerating inside and copying outside, inherit every one of these errors.
We present HintErase, which treats localization as a hint, not a hole.
The model sees the full, unmasked video together with a hint channel; during training the hint is corrupted the way real localizers fail, while supervision follows the true overlay.
The result is a single model that removes overlay the hint missed, keeps background the hint wrongly covered, and is trained to keep genuine scene text even when the hint lands on it.
For full-length releases, a streaming pipeline handles interlaced sources, shot cuts, missed and jittering hints, paste-back and audio, with memory independent of the length of the video.
This report describes the approach, the pipeline that turns it into full-length deliveries, and qualitative results; video results, with input, hint and output side by side, are on the project page.
\end{abstract}

\section{Introduction}
\label{sec:intro}

Overlays are composited onto video after capture: hard subtitles, karaoke lyrics, title cards, end credits, station logos and watermarks.
Removing them is a prerequisite for re-releasing film libraries, series and music videos in new languages and on new platforms.
The task is demanding in ways that object removal is not.
Overlays are thin, outlined, glowing or semi-transparent; they appear, change color and vanish with the soundtrack; and one release carries thousands of overlaid frames that must all be erased without a visible seam.

Diffusion-based video models have made the \emph{filling} half of the problem largely tractable~\cite{li2025diffueraser,zi2025minimaxremover,miao2026deoverlay}.
Running such models on complete releases, we found that most remaining failures traced back to \emph{localization} instead.
Residue survived wherever the segmenter missed an outline or a glow, wherever the detector skipped a frame, and on small elements it never detected.
Dilating the mask to catch these cases made the model repaint clean background between characters, which blurred texture and introduced flicker.
The bottleneck had moved from how to fill to where to fill.

Existing methods resolve the question in one of two ways.
\emph{Mask-as-hole} erasers~\cite{zhou2023propainter,li2025diffueraser,zi2025minimaxremover} regenerate the masked region and copy everything else, so any overlay pixel outside the mask survives, whatever the generator (\cref{sec:formulation}).
\emph{Mask-free} subtitle removers~\cite{he2026clear,hui2026sedit} need no mask, but cannot be told where to act and may erase genuine text in the scene.
Both assume that the mask is either exact or absent.

We argue that localization should be read as a noisy observation of where overlays are, a \emph{hint}, and not as the definition of where to generate, a \emph{hole}.
A remover built this way should be \emph{complete}, removing overlay that the hint misses; \emph{faithful}, keeping clean pixels that the hint covers; and \emph{selective}, keeping genuine scene text and other overlay-like scene content even where the hint covers it.

Our contributions are:
\begin{itemize}
  \item \ours, a video overlay remover that conditions on the full video and a hint channel, and is trained with hints corrupted like real localizer output under supervision that follows the true overlay. One model accepts exact masks, loose boxes, partial masks or only a coarse region (\cref{sec:method}).
  \item A streaming pipeline that turns the model into full-length deliveries: deinterlacing, shot segmentation, temporal hint repair, windowed inference with fixed memory, brightness-adaptive paste-back and frame-exact output with the original audio (\cref{sec:delivery}).
  \item Qualitative results on karaoke lyrics, title cards with logos, theme-song lyrics and end credits in animation, imperfect hints and a multi-shot music video (\cref{sec:results}); the video results play on the project page with input, hint and output side by side.
\end{itemize}

\section{Related Work}
\label{sec:related}

\paragraph{Video inpainting and object removal.}
Propagation-based inpainters fill a mask by transferring visible content along time~\cite{zeng2020sttn,li2022e2fgvi,zhou2023propainter}.
Diffusion models add a generative prior for content that is never revealed~\cite{zhang2024avid,li2025diffueraser}, and diffusion transformers now lead object removal~\cite{bian2025videopainter,jiang2025vace,liu2025eraserdit,zi2025minimaxremover}.
These methods define the region to regenerate by the mask, which is the mask-as-hole design analyzed in \cref{sec:formulation}.

\paragraph{Overlay and text removal.}
Image scene-text removal~\cite{liu2020erasenet,liu2022ctrnet,peng2024viteraser} and visible watermark removal~\cite{liu2021wdnet,cun2021splitnet,chen2025deco} are mature for single images or a single overlay type.
For video subtitles, CLEAR~\cite{he2026clear} and SEDiT~\cite{hui2026sedit} remove text without a mask, the latter in one sampling step with chunk-wise streaming for long videos.
DeOverlay~\cite{miao2026deoverlay} starts a flow from the source latent, aggregates visible source observations over time inside each window, and introduces DVTE-Bench with eight text-overlay categories.
Like DeOverlay, we keep the source visible.
Our focus is different: how removal should behave when the mask is wrong, and how a full release should be processed.

\paragraph{Long videos.}
Training-free long-video methods co-denoise overlapping windows and average them~\cite{wang2023genlvideo,lyu2025unified}, shift a queue of frames at increasing noise levels~\cite{kim2024fifo}, or condition each chunk on the previous one~\cite{henschel2025streamingt2v}.
We process full releases in a streaming pipeline whose memory is independent of the length of the video (\cref{sec:delivery}).

\section{Hint, Not Hole}
\label{sec:formulation}

Let $\src$ be a video with overlays and $\tgt$ its clean counterpart.
The \emph{true} overlay region $\G$ contains every pixel an overlay affects, including glow, shadow and compression halo.
A localizer returns a \emph{hint} $\Hm$: a segmentation mask, a box, a user stroke, or only a coarse region.
Real hints err in two directions, \emph{under-coverage} $\G\setminus\Hm$ (missed strokes, halos, whole frames and whole objects) and \emph{over-coverage} $\Hm\setminus\G$ (dilation margins, box corners, gaps between characters, false positives).
A remover $f(\src,\Hm)$ is \emph{complete} if $f\approx\tgt$ on all of $\G$, \emph{faithful} if $f\approx\src$ outside $\G$, and \emph{selective} if it stays faithful on scene content that resembles an overlay, such as scene text, even where $\Hm$ covers it.
Selectivity is the hardest case of faithfulness, because that content looks like what is being removed.

Mask-as-hole erasers keep the input outside the mask by construction, through blanked conditioning, latent blending or paste-back.
That alone sets a floor on what survives.

\begin{proposition}[Residual floor]
\label{prop:floor}
If $f(\src,\Hm)=\src$ outside $\Hm$, then on $\G\setminus\Hm$ the output equals the input, so the residual share
\begin{equation}
  \frac{\sum_{\G}|f(\src,\Hm)-\tgt|}{\sum_{\G}|\src-\tgt|}\;\ge\;\rho(\G,\Hm)=\frac{\sum_{\G\setminus\Hm}|\src-\tgt|}{\sum_{\G}|\src-\tgt|}
\end{equation}
for every generator.
\end{proposition}
Dilating $\Hm$ lowers $\rho$ only by enlarging $\Hm\setminus\G$, where the same erasers discard the input and must invent background they could have copied.
Completeness and faithfulness therefore trade off by construction.
Breaking the trade-off needs a model that may edit outside the hint and may copy inside it, \ie one that reads the hint as evidence.

\begin{figure}[t]
\centering
\setlength{\tabcolsep}{1pt}\renewcommand{\arraystretch}{0.6}
\begin{tabular}{@{}p{0.30\linewidth} p{0.30\linewidth} p{0.37\linewidth}@{}}
 \cell[0.8]{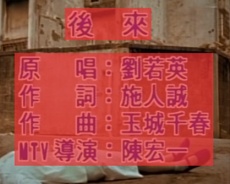} & \cell[0.8]{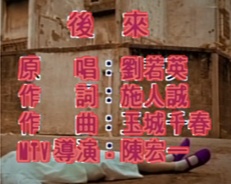} & \cell[0.622]{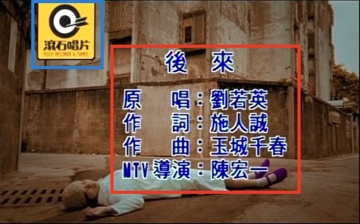} \\
 \centering\collab{(a) Region mask} & \centering\collab{(b) Character mask} & \centering\arraybackslash\collab{(c) Box hint (\ours)} \\
\end{tabular}
\caption{\textbf{Three ways to localize the same title card.} (a) and (b) are too large and too small when read as a hole; (c) is coarser than both, yet sufficient when read as a hint. Red: text; blue: logo.}
\label{fig:why}
\end{figure}

\Cref{fig:why} shows the trade-off on one title card.
A region mask (a) covers whole lines, so a hole-filling eraser discards the clean background between and around the characters and has to repaint it.
A character-level mask (b) follows the strokes, but outlines, glows and compression halos fall just outside it and survive, together with any stroke the segmenter missed.
Read as a hint, one box per text block and per logo (c) is enough: the model sees the frame with the overlay in place, removes everything that belongs to the overlay, and copies the background inside the box.

\section{\ours}
\label{sec:method}

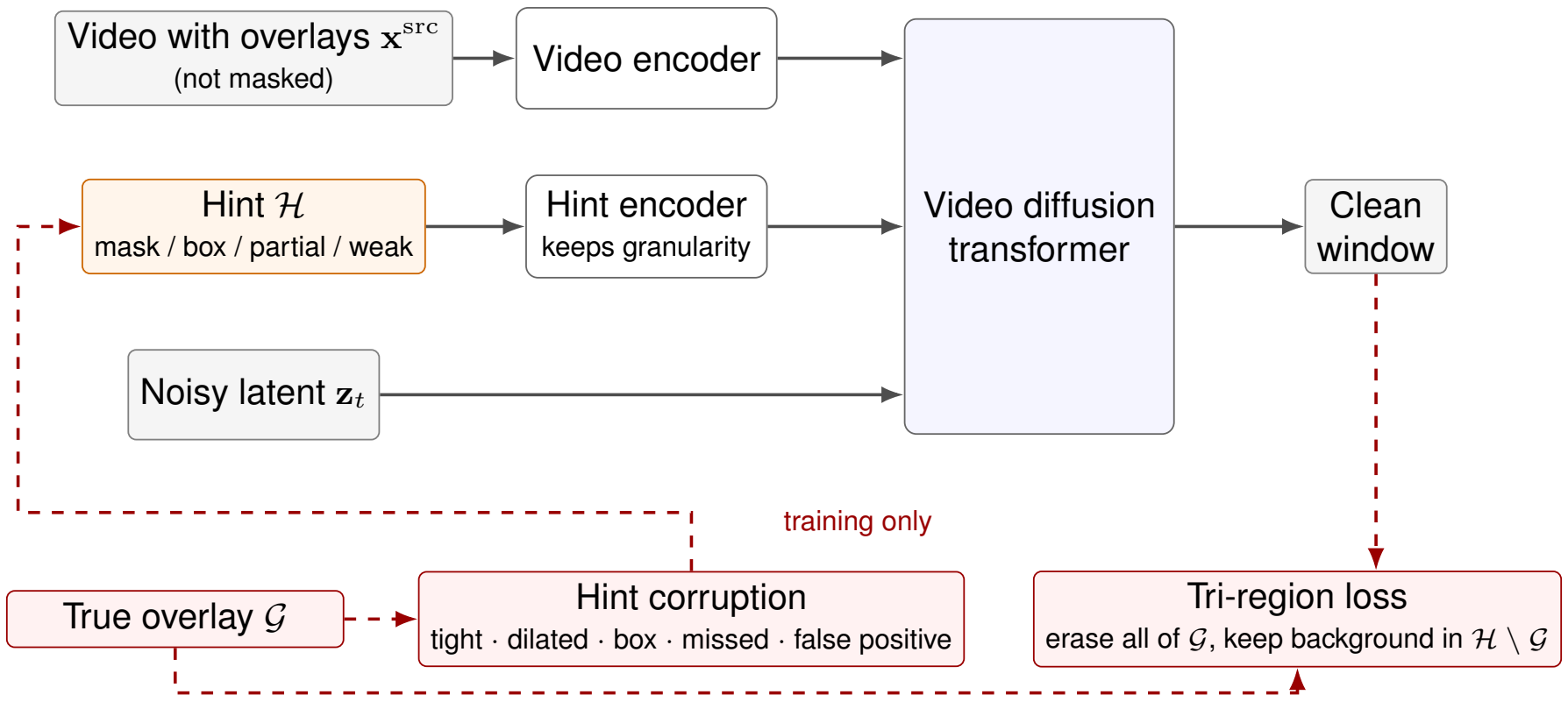
\begin{figure*}[t]
\centering
\resizebox{0.66\textwidth}{!}{%
\begin{tikzpicture}[
  font=\sffamily\small,
  box/.style={draw=black!60, rounded corners=3pt, align=center, minimum height=0.9cm, inner sep=4pt, fill=white},
  data/.style={draw=black!50, fill=black!4, rounded corners=2pt, align=center, minimum height=0.8cm, inner sep=3pt},
  hint/.style={draw=orange!80!black, fill=orange!8, rounded corners=2pt, align=center, minimum height=0.8cm, inner sep=3pt},
  train/.style={draw=red!60!black, fill=red!5, rounded corners=2pt, align=center, inner sep=3pt},
  arr/.style={-{Latex[length=2mm]}, thick, black!70},
  darr/.style={-{Latex[length=2mm]}, thick, dashed, red!60!black},
]
\node[data] (src) at (0,2.0) {Video with overlays $\src$\\[-1pt]\scriptsize (not masked)};
\node[hint] (hint) at (0,0.5) {Hint $\Hm$\\[-1pt]\scriptsize mask / box / partial / weak};
\node[data] (noise) at (0,-1.0) {Noisy latent $\mathbf{z}_t$};
\node[box] (vae) at (3.5,2.0) {Video encoder};
\node[box] (fold) at (3.5,0.5) {Hint encoder\\[-1pt]\scriptsize keeps granularity};
\node[box, fill=blue!4, minimum height=3.7cm, minimum width=2.4cm] (dit) at (7.0,0.5) {Video diffusion\\transformer};
\node[data] (out) at (10.0,0.5) {Clean\\window};
\draw[arr] (src) -- (vae); \draw[arr] (hint) -- (fold);
\draw[arr] (vae.east) -- (dit.west |- vae.east);
\draw[arr] (fold.east) -- (dit.west |- fold.east);
\draw[arr] (noise.east) -- (dit.west |- noise.east);
\draw[arr] (dit) -- (out);
\node[train, minimum width=3.0cm] (gt) at (-0.7,-3.0) {True overlay $\G$};
\node[train, minimum width=3.4cm] (deg) at (3.9,-3.0) {Hint corruption\\[-1pt]\scriptsize tight $\cdot$ dilated $\cdot$ box $\cdot$ missed $\cdot$ false positive};
\node[train, minimum width=3.4cm] (loss) at (9.3,-3.0) {Tri-region loss\\[-1pt]\scriptsize erase all of $\G$, keep background in $\Hm\setminus\G$};
\draw[darr] (gt) -- (deg);
\draw[darr] (deg.north) |- (-2.1,-2.05) |- (hint.west);
\draw[darr] (gt.south) -- ++(0,-0.4) -| (loss.south);
\draw[darr] (out.south) -- (out.south |- loss.north);
\node[anchor=west, text=red!60!black, font=\sffamily\scriptsize] at (4.6,-2.15) {training only};
\end{tikzpicture}}
\caption{\textbf{Overview of \ours.} The unmasked video and a separate hint channel condition a video diffusion transformer. In training, the hint is a corrupted copy of the true overlay region $\G$, while the loss is defined by $\G$ itself and by background the hint wrongly covers, so supervision never trusts the hint.}
\label{fig:overview}
\end{figure*}

\ours{} fine-tunes a pretrained video diffusion transformer trained with flow matching~\cite{lipman2023flow,liu2023rectified} in a VAE latent space.
\Cref{fig:overview} shows a single window.

\paragraph{Hint as observation.}
The model always sees the full, unmasked video.
Its input concatenates the noisy latent, the latent of the source video and an encoding of the hint,
\begin{equation}
  \mathbf{z}_{\text{in}} = \big[\,\mathbf{z}_t \;\|\; \mathcal{E}(\src) \;\|\; \phi(\Hm)\,\big],
  \label{eq:input}
\end{equation}
so the overlay pixels the hint misses and the clean pixels it covers both reach every layer.
The hint encoder $\phi$ preserves how much of each latent cell the hint covers, so a tight character mask and a loose box over the same line stay distinguishable after downsampling.
This tells the model what kind of localizer it is dealing with, and how far to trust it.
With a weak hint, such as a coarse region instead of a mask, the same weights still remove the overlays inside it, so they do not depend on an exact mask.

\paragraph{Training with corrupted hints.}
Robustness is only learned for errors seen in training.
The hint given to the model is therefore a corrupted copy of the true region, drawn from a model of the errors real front-ends make: tight masks, over-dilated masks, boxes, masks with missing strokes, frames or whole objects, and false positives, some of them placed on genuine scene text.
Training clips are clean videos with overlays synthesized on the fly, so the clean target is known exactly everywhere.

\paragraph{Tri-region objective.}
Corrupting the input is not enough.
If supervision followed the hint, the model would learn to regenerate exactly the hinted area and nothing else, which is the trade-off of \cref{prop:floor} again.
We weight the flow-matching loss by regions defined on the \emph{truth},
\begin{equation}
  w = 1 + \lambda_G\, \tilde{G} + \lambda_H\, \max\!\big(H - \tilde{G},\, 0\big),
  \label{eq:triloss}
\end{equation}
where $H$ is $\Hm$ on the latent grid and $\tilde G$ is $\G$ on the latent grid, slightly dilated to include ringing.
The overlay term covers all of $\G$ whether or not the hint does, which pays for completeness exactly where mask-as-hole erasers are blind.
The hinted-background term covers $\Hm\setminus\G$, where the hint invites editing but the target equals the input, which pays for faithfulness.
When a false-positive hint lands on scene text, the same term teaches selectivity.

\subsection{From model to delivery}
\label{sec:delivery}

The model is trained and evaluated on clips of a few seconds; a delivery is a complete video of thousands of frames, and three gaps open between the two.
\emph{Short window, long video}: the video has to be split into windows and reassembled without seams, under a fixed memory budget.
\emph{One shot, frequent cuts}: a window that crosses a cut lets the shots on either side bleed into each other, while fades, flashes and dissolves look like cuts and are not.
\emph{Exact masks, imperfect localizers}: a frame the detector misses keeps its text, and a mask that jitters from frame to frame makes the output flicker.
None of these shows on a short test clip; they surface together on a full-length video, and a failure in any one of them makes the delivery unusable.
\Cref{tab:delivery} lists the pipeline stage by stage, as a general method independent of any particular tuning; the specific values used to produce every result in this report are given separately below the table.

\begin{table*}[!t]
\centering\footnotesize
\setlength{\tabcolsep}{5pt}\renewcommand{\arraystretch}{1.1}
\begin{tabular}{@{}p{0.12\textwidth} p{0.30\textwidth} p{0.52\textwidth}@{}}
\toprule
Stage & Problem & Solution \\
\midrule
Input & Sources are interlaced, heavily compressed, variable-frame-rate, or stored with non-square pixels. Small text in low-resolution sources covers only a few latent cells. &
Interlacing is detected by a majority vote over an initial window of decoded frames and removed with a field-parity-aware deinterlacer; the sample aspect ratio and frame timing are carried through to the output. Each run is cropped around the union of its hints and resized to a fixed working size chosen for the overlay type, so small text is magnified before encoding. \\
Shots & Windows that cross a cut mix two shots; fades, flashes and dissolves trigger false cuts; detectors leave fragments of a few frames. &
A shot-boundary detector proposes cuts with a probability. Confident cuts are kept; the others are kept only if the frames on either side differ in both a gray thumbnail and a color histogram, with a looser test for low-probability cuts. Shots shorter than a minimum length are merged into a neighbor. Runs, windows and temporal hint extension never cross a remaining cut. \\
Hints & Segmenters miss frames, especially during fades; masks jitter; small marks such as logos and title cards are missed by a text segmenter. &
The character-level subtitle mask is combined with boxes from a logo and title-card detector. Missed frames are filled in time: a gap between two hinted frames, up to a bounded length, is filled with the union of the hints within a fixed radius on either side, and a hint is extended by a bounded number of frames before an overlay first appears and after it last disappears. Hints need no spatial dilation, since the model reads them as evidence. \\
Planning & A full video does not fit in GPU memory; most frames carry no overlay at all. &
A first decoding pass records frame times, thumbnails and per-frame hint statistics. Frames with a hint are grouped into runs, padded with context frames and merged when close enough together; only runs enter the model. Each run is covered by overlapping windows of fixed length. A second pass streams the video, and each frame is written as soon as no later window needs it; memory is bounded by the window length, independent of the length of the video. \\
Paste-back & Replacing whole windows softens untouched areas; residual text on dark frames differs from the background by only a few gray levels. &
Only pixels the model changed are pasted back: a pixel is replaced if it differs from the input by more than an adaptive threshold that drops with the mean luma of the frame, so fades and night scenes are not under-replaced. The replaced region is restricted to a fixed neighborhood of the hint at the working resolution and feathered at its border. \\
Output & Dropped or duplicated frames; audio drifts out of sync. &
Frames outside runs are written without entering the model, and the output has exactly as many frames as the source. The source audio is muxed back, compensated for the offset between the video stream and the container start, and trimmed to the video duration; truncating to the shorter stream instead can drop the last frame. \\
Ultra-HD & Old footage is low-resolution, while re-releases often require Ultra-HD; generative upscalers built on large diffusion models~\cite{yao2026flashvsrplus} can shift colors, alter faces and people, and invent details. &
Optional: after erasure, the clean video is upscaled by a fixed factor with an in-house super-resolution model designed to preserve the original colors, people and details. \\
\bottomrule
\end{tabular}
\caption{\textbf{From model to delivery.} Each stage of the long-video pipeline, the problem it addresses on full-length releases, and how it is solved in general. Cost scales with the duration of the overlays, not with the length of the video.}
\label{tab:delivery}
\end{table*}

{\footnotesize\noindent\textit{Values used for every result in this report:} deinterlacing vote over the first 300 decoded frames; working size $448\times1280$ for subtitle bands and $576\times1024$ otherwise; shots of four frames or fewer merged into a neighbor; hint gaps of up to twelve frames bridged by a union within six frames on either side, hints extended by up to three frames at onset and offset; runs padded by eight frames of context and merged when at most 24 frames apart; 81-frame windows; paste-back threshold $\tau=\max\!\big(3,\ 12\cdot\min(1,\bar{\ell}/96)\big)$, where $\bar{\ell}$ is the mean luma of the frame, applied within a 24-pixel neighborhood of the hint; $4\times$ upscaling for Ultra-HD delivery.}

The pipeline needs no manual annotation: once the front-ends have produced hints, a release is processed end to end in a single run.

\section{Results}
\label{sec:results}

\begin{figure*}[t]
\centering
\setlength{\tabcolsep}{1pt}\renewcommand{\arraystretch}{0.6}
\begin{tabular}{@{}c*{5}{p{0.1925\textwidth}}@{}}
 & \centering\collab{(a) Character mask} & \centering\collab{(b) Mask eroded by 2\,px} & \centering\collab{(c) Mask missing strokes} & \centering\collab{(d) Filled box per line} & \centering\arraybackslash\collab{(e) Detection box} \\
\rowlab{Hint} & \cell[0.35]{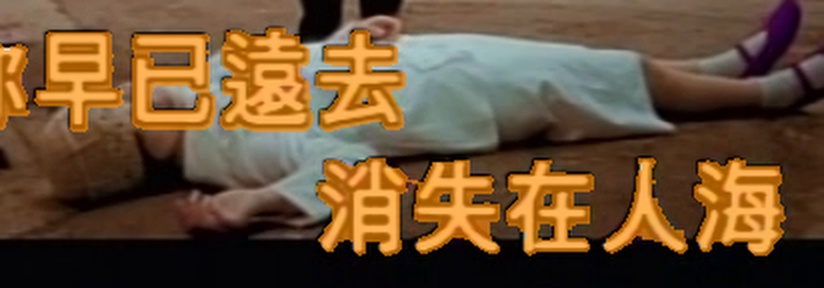} & \cell[0.35]{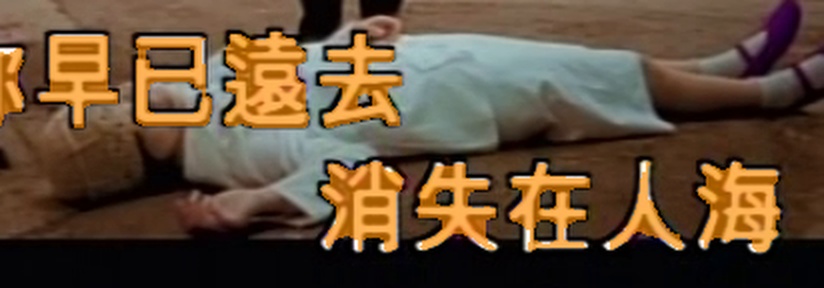} & \cell[0.35]{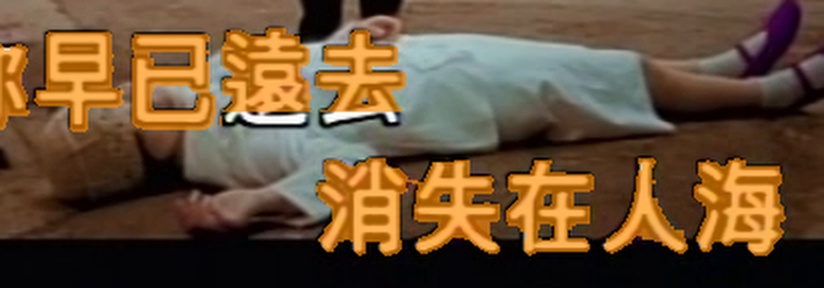} & \cell[0.35]{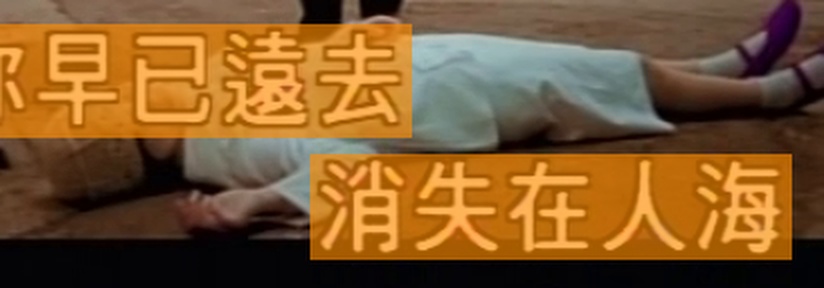} & \cell[0.35]{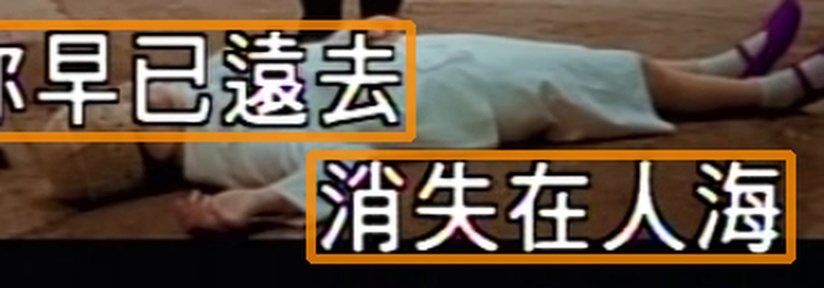} \\
\rowlab{Output} & \cell[0.35]{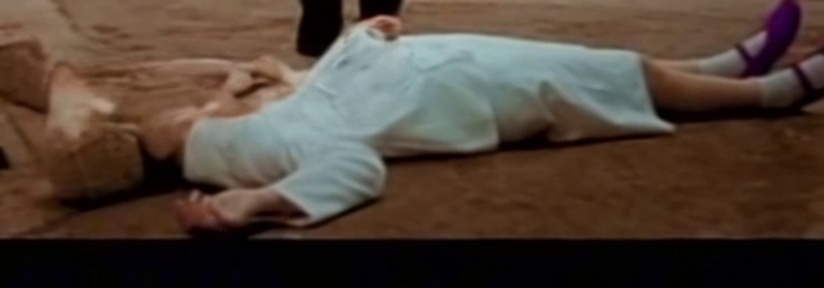} & \cell[0.35]{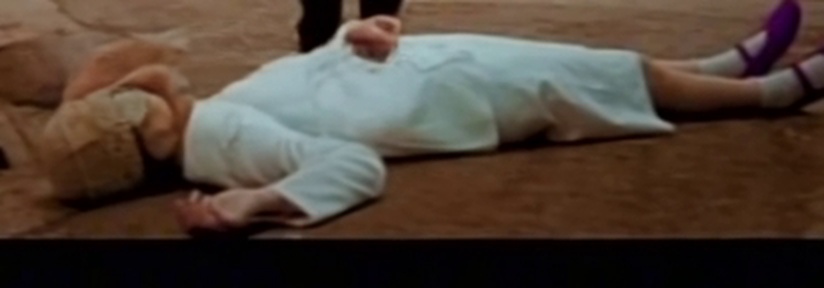} & \cell[0.35]{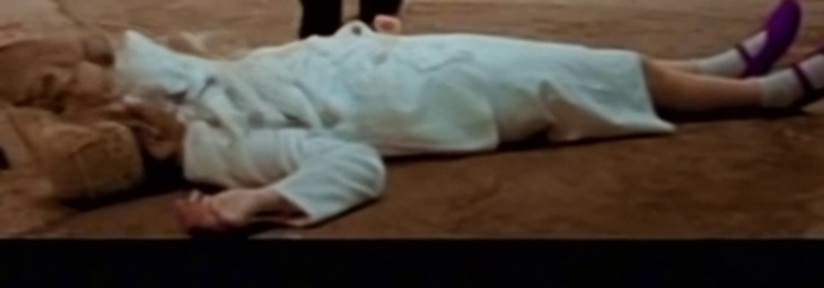} & \cell[0.35]{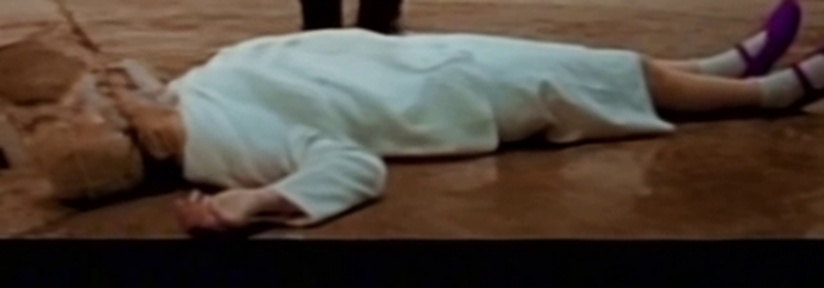} & \cell[0.35]{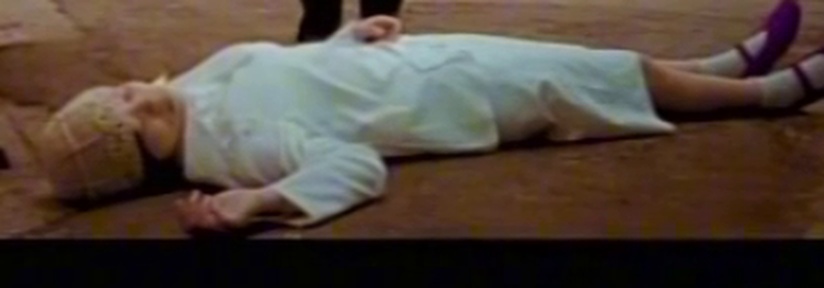} \\
\end{tabular}
\caption{\textbf{The same weights under five hints.} Top: the hint (orange) on the input. Bottom: the output of \ours{}, which reads the hint as evidence. A mask-as-hole eraser cannot change pixels outside the hint (\cref{prop:floor}), so it would leave the outlines just outside the character mask (a), the halo outside the eroded mask (b), the lower strokes of the last two characters (c), and all of the lyrics inside a detection box drawn as its outline (e); with a filled box (d) it would have to regenerate all the clean background inside the box. \ours{} removes the lyrics under every hint.}
\label{fig:hints}
\end{figure*}

\begin{figure}[!t]
\centering
\setlength{\tabcolsep}{1pt}\renewcommand{\arraystretch}{0.8}
\begin{tabular}{@{}c m{0.82\linewidth}@{}}
 \rowlab{Input} & \cell[0.39]{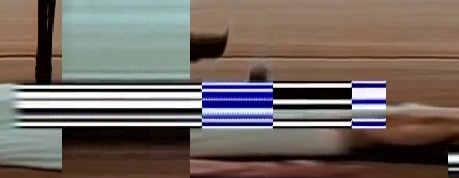} \\
 \rowlab{\ours} & \cell[0.39]{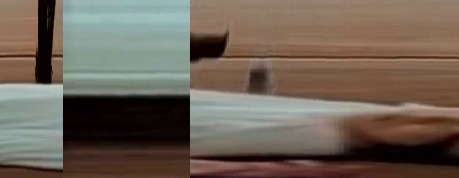} \\
\end{tabular}
\caption{\textbf{Time slice through the lyrics band.} Each image is one pixel column through the lyrics of the karaoke clip that opens the project page, with 18 seconds of frames running left to right. In the input, lyric lines appear as striped blocks that change color as the wipe passes; in the output, the streaks of the background run through them without vertical steps.}
\label{fig:slices}
\end{figure}

\paragraph{Setup.}
Hints come from the production front-ends: a character-level subtitle segmenter and a logo and title-card detector.
All masks and boxes in this report are computed by these models; \cref{fig:hints} additionally perturbs them, as described below. Except for the hint comparison in \cref{fig:hints}, the hint is drawn as one box per text block or logo, in the figures and on the project page.
All clips come from licensed productions and are shown with the permission of the rights holders.
None of them were used in training; the training data was filtered to exclude them.
We show \ours{} only. The question this report asks is how removal should behave when the hint is wrong, and we answer it by running the same weights with different hints (\cref{fig:hints}).
Still frames cannot show flicker or seams; the clips in \cref{fig:teaser} play on the project page, where input, hint and output run side by side on a shared timeline.

\paragraph{Overlay types.}
\Cref{fig:teaser} shows one example for each of four overlay types.
Karaoke lyrics are among the most demanding subtitle cases in our footage, because the fill changes color mid-line under an outline that does not, and countdown dots appear before each line (\pagecase{ktv-lyrics}).
Title cards stack several lines of credits next to a label logo, and fade in and out with the shot (\pagecase{ktv-title}); the logo is hinted by its own box.
Theme-song lyrics in animation sit on top of fast, saturated motion (\pagecase{anime-lyrics}), and credit lines at the end of an episode mix fonts and colors over busy scenes (\pagecase{anime-credits}).
Low-resolution animation has flat colors and thin outlined text, which make any residue easy to see.

\paragraph{Imperfect hints.}
\Cref{fig:hints} runs the same weights on one clip under five hints that a real localizer could produce: a character mask, the same mask eroded so that it misses the outlines, the same mask missing the lower 20\% of the strokes of two characters where they cross a white garment, a filled box per line, and the outline of a detection box.
Read as a hole, each of them fails in a predictable way: by \cref{prop:floor}, nothing outside the hint can change, so the dark outlines just outside even the character mask, the halos outside the eroded mask and the uncovered strokes all remain, and a detection box drawn as its outline would regenerate the outline and nothing else; a filled box covers the lyrics but would discard the clean background inside it.
Read as a hint, the same pixels are visible to the model, and \ours{} removes the lyrics under all five hints, so the localizer can be as coarse as a box per line, as in \cref{fig:why}(c).

\paragraph{Full-length videos.}
\Cref{fig:slices} looks at the karaoke clip that opens the project page through a time slice: one pixel column through the lyrics band, with frames running left to right.
A static background appears as horizontal streaks, a shot cut as a vertical edge, and a seam or flicker between windows as a vertical step where no cut is.
The clip is cut from a complete music video processed in one run with the pipeline of \cref{sec:delivery}, and the slice shows the lyrics band continuing the background across shot after shot without steps.

\paragraph{Ultra-HD delivery.}
Old karaoke footage is often re-released in Ultra-HD, and upscaling is where a re-release most easily loses fidelity: generative video upscalers built on large diffusion models, such as FlashVSR+~\cite{yao2026flashvsrplus}, can shift colors, alter faces and people, and invent texture. The pipeline therefore erases the overlays at the source resolution and then upscales the clean video $4\times$ with an in-house super-resolution model designed for fidelity rather than generation (\cref{tab:delivery}). Upscaling also magnifies whatever the eraser leaves behind, so the erased regions have to stay consistent with their surroundings at the delivered resolution.

\section{Conclusion}
\label{sec:conclusion}

\paragraph{Conclusion.}
Generative models have made filling a known region the easy part of overlay removal; deciding what the region is remains hard, and practical localizers err in both directions.
\ours{} treats the localizer's output as a hint: it observes the full video, learns from hints corrupted like real ones, and is supervised by the truth.
For full releases, a streaming pipeline handles deinterlacing, shot cuts, imperfect hints, paste-back and audio, so a complete video is processed in one run.
The model does not need a better localizer, only a plausible one.

\paragraph{Broader impact.}
Overlay removal can be misused to strip watermarks or attribution from content one has no right to modify.
\ours{} was developed to re-release licensed content; deployments should keep provenance metadata and restrict watermark removal to rights holders.

{\small
\bibliographystyle{unsrtnat}
\bibliography{refs}
}

\end{document}